# Improved Accuracy of PSO and DE using Normalization: an Application to Stock Price Prediction

Savinderjit Kaur
Department of Information Technology, UIET, PU,
Chandigarh, India

Veenu Mangat
Department of Information Technology, UIET, PU,
Chandigarh, India

*Abstract*— **Data Mining is being actively applied to stock market since 1980s. It has been used to predict stock prices, stock indexes, for portfolio management, trend detection and for developing recommender systems. The various algorithms which have been used for the same include ANN, SVM, ARIMA, GARCH etc. Different hybrid models have been developed by combining these algorithms with other algorithms like roughest, fuzzy logic, GA, PSO, DE, ACO etc. to improve the efficiency. This paper proposes DE-SVM model (Differential Evolution-Support vector Machine) for stock price prediction. DE has been used to select best free parameters combination for SVM to improve results. The paper also compares the results of prediction with the outputs of SVM alone and PSO-SVM model (Particle Swarm Optimization). The effect of normalization of data on the accuracy of prediction has also been studied.**

## I. INTRODUCTION

Stock Market prediction is an attractive field for research due to its commercial applications and the attractive benefits it offers. It follows stochastic, non-parametric and nonlinear behavior. An important hypothesis related to stock market which has been debated and researched time and again is EMH (Efficient Market Hypothesis). According to EMH, the stock market immediately reflects all of the information available publicly. But in reality, the stock market is not that efficient, so the prediction of stock market is possible.

This paper proposes a hybrid of DE-SVM (Differential Evolution-Support Vector Machines). The performance of SVM is based on the selection of free parameters C (cost penalty), $\epsilon$ (insensitive-loss function) and $\gamma$ (kernel parameter). DE will be used to find the best parameter combination for SVM. DE-SVM has already been used by Zhonghai Chen et al. [6] for air conditioning load prediction, Yong Sun et al. [7] for gas load prediction, Jośe Garćıa-Nieto et al. [8] for feature selection, Shu Jun et al. [9] for rainstorm forecasting and for studying the lithology identification method from well logs by Jiang An-nan et al. [10]. The paper also compares the results of DE-SVM with PSO-SVM and SVM. The effect of normalization on datasets has also been studied.

## II. LITERATURE REVIEW

Yohanes et al. [1] showed that ARIMA (Autoregressive Integrated Moving Average) can be outperformed by ANN. ESS (Each sum square) result with ARIMA is 284.95 and with ANN is 170.40 [1]. Qiang Ye et al. [2] proved that stock price prediction results using amnestic NN are better than common ANN. The ratio of right classified stocks is 58.25% when forgetting coefficient is 0.10 as compared to 56.25% for forgetting coefficient of 0.00 (for common ANN) [2]. Ling-Feng Hsieh et al. [3] integrated DOE (Design of Experiment) with BPNN to show that experimental validation of the optimal parameter settings can effectively improve the forecasting rate to 84%. Mustafa E. Abdual-Salam et al. [4] proved that DE converges to global minimum faster and gives better accuracy than PSO when used as training algorithms for ANN. Zhang Da-yong et al. [5] proposed a hybrid model ARMA-SVM (Autoregressive Moving average-SVM) which has MSE of 1.1433 against 1.1494 for BPNN.

### A. *Support Vector Machines (SVM):*

SVM was developed by Vapnik and Cortes in 1995. SVM is a promising method for the classification of both linear and nonlinear data [11]. SVM can be used both for classification and regression. SVMs can be trained with lesser input samples and are less prone to overfitting. The training time of even the fastest SVMs can be extremely slow, but they are highly accurate, owing to their ability to model complex nonlinear decision boundaries [11]. SVM follows supervised learning. For classification purposes, when data is linearly separable a straight line can be drawn to separate the tuples of one class from the other. For nonlinear data, the data is mapped into higher dimensional space where the different classes can be separated using a hyperplane. A number of hyperplanes are possible but SVM searches for the maximum marginal hyperplane (MMH). The vectors in the training set that have minimal distance to the maximum margin hyperplane are called support vectors [12].

SVM selects the minority of observations (support vectors) to represent the majority of the rest of the observations [13]. The soft margins were introduced to penalize but not prohibit classification errors while finding the maximum margin hyperplane [11].

If the margin can be significantly increased, the better generalization can outweigh the penalty for a classification error on the training set [11]. To maximize the prediction ability of a model, both underfitting and overfitting need to be depressed at the same time in data processing [25]. The error of training is called Empirical Risk denoted by $R_{emp.}$ SVM uses SRM (Structural Risk Minimization) instead of ERM (Empirical Risk Minimization) which aims at minimizing (1) :

$$\min \left[ R_{emp+} \sqrt{\frac{h\left(\ln\frac{2l}{h}+1\right)-\ln\left(\frac{\eta}{4}\right)}{4}} \right] \quad (1)$$

Here, l is number of samples in training set, 1-η is the probability of the equation ( (1) ≥ $R_{pred}$ , $R_{pred}$ is the total risk of prediction) to be true and h is VC dimension to depress overfitting in data processing [25].

*SVM parameters:* The performance of SVM is based on three basic parameters C (cost penalty), ϵ (insensitive loss function parameter) and γ (kernel parameter).

Cost penalty: C determines the trade-off cost between minimizing the training error and minimizing the model's complexity [26]. The parameter C determines the trade-off between model complexity and the tolerance degree of deviations larger than ε [20].

ϵ loss-insensitive function: Parameter ϵ controls the width of the ϵ-insensitive zone, used to fit the training data [27]. Larger ϵ-value result in fewer SVs selected, and result in more 'flat'(less complex) regression estimates [20]. If the value of ϵ is too big, the separating error is high, the number of support vectors is small, and vice versa [26].

Kernel parameter: γ ($2\sigma^2$) of the kernel function implicitly defines the nonlinear mapping from input space to some high-dimensional feature space [28]. The main kernels used are:

1) Linear kernel: x.y

2) Polynomial kernel: $K(x_i,x_j)= (x_i.x_j+1)^d$

3) Radial Basis kernel: $K(x_i,x)=\exp(-\frac{\|x_i-x\|^2}{2\sigma^2})$

4) Sigmoid kernel function: $K(x_i,x_j)=\tanh(x_i.x_j+p)$

RBF kernel is mostly used for stock price prediction because only one parameter needs to be confirmed, there are less SVR training parameters constructed by it and it is easy to confirm SVR training parameters [18]. The kernel width parameter σ in RBF is appropriately selected to reflect the input range of the training/test data. For univariate problems, RBF width parameter is set to σ ~[0.1–0.5]* range(x) [20].

### *B. Differential Evolution (DE):*

Differential evolution (DE) was introduced by Kenneth Price and Rainer Storn in 1995 for global continuous optimization problem. It has won the third place at the 1st International Contest on Evolutionary Computation [14]. DE belongs to the family of Evolutionary Algorithms (EA). DE algorithm is similar to genetic algorithms having similar operations of crossover, mutation and selection. DE can find the true global minimum regardless of the initial parameter values. DE provides fast convergence and uses fewer control parameters. DE constructs better solutions than genetic algorithms because GA relies on crossover while DE relies on mutation operation. It is a stochastic population-based search method that employs repeated cycles of recombination and selection to guide the population towards the vicinity of global optimum. DE uses a differential mutation operation based on the distribution of parent solutions in the current population, coupled with recombination with a predetermined parent to generate a trial vector (offspring) followed by a one-to-one greedy selection scheme between the trial vector and the parent [15]. Depending on the way trial vector is generated, there exist many trial vector generation strategies and consequently many DE variants. High convergence characteristics and robustness of DE have made it one of the popular techniques for real-valued parameter optimization. DE uses three parameters conventionally, they are: the population size NP, the scale factor F and the crossover probability CR/ Cr. Some conditions for these variables include: NP>4, F>0 and is a real valued constant and is often set to 0.5, CR Є (0, 1) and is often set to 0.9 [16]. Different stages in DE are:

1. Population structure : The current population, symbolized by $P_c$, is composed of those D-dimensional vectors $X^g_i = \{x^g_{i,1}, x^g_{i,2}, \ldots, x^g_{i,D}\}$, the index g indicates the generation to which a vector belongs [17]. In addition, each vector is assigned a population index, i, which varies from 1 to $N_p$, knowing that $N_p$ is the population size [17]. Once initialized, DE mutates randomly chosen vectors to produce an intermediary population $P_v$ of Np mutant vectors $V_i^g$ [22]. Each vector in the current population is then recombined with a mutant to produce a trial population $P_u$ of $N_p$ trial vectors $U_i^g$ [22].

2. Initialization : This stage consists in forming the initial population. For example, if our objective is the optimization of the membership functions, the initialization step consists in arbitrarily choosing the interval of this function [17].

3. Mutation [17, 22]: For each vector (for example, a vector which represents the interval of the membership functions) $V_i^g=\{v^g_{i,1}, v^g_{i,2},\ldots, v^g_{i,D}\}$ a mutant vector is produced according to the following formulation [22]:

$$v^g_{i,j}=x^g_{r0,j} + F(x^g_{r1,j} - x^g_{r2,j}) \quad (2)$$

The scale factor F is a positive real number that controls the rate at which the population evolves. While there is no upper limit on F, effective values seldom are greater than 1.

4. Crossing [17,22,4]:The relative vector is mixed with the transferred vector to produce a test vector $T^g_{i,j}$:

$$T^g_{i,j}=\begin{cases} v^g_{j,i} & \text{if } (r^g_{j,i} \le CR \text{ or } j=j_r) \\ x^g_{j,i} & \text{otherwise} \end{cases} \quad (3)$$

The crossover probability CR ϵ [0,1] is a user-defined value that controls the fraction of parameter values that are copied from the mutant. To determine which source contributes, a given uniform crossover parameter compares CR to the output of a uniform random number generator $r_{j,i}^{g}$. If the random number is less than or equal to CR, the trial parameter is inherited from the mutant $v_{j,i}^{g}$ ; otherwise, the parameter is copied from the vector $x_{j,i}^{g}$ . In addition, the trial parameter, with randomly chosen index $j_r$ is taken from the mutant to ensure that the trial vector does not duplicate $x_{i}^{g}$. Because of this additional demand, CR only approximates the true probability.

5. Selection [17]: All the solutions in the population have the same chance that the parents of being selected, regardless of their fitness function value. The child produced (new vector) after the crossing operations is evaluated. Then, the performances of the child vector and its relative are compared and the best one is selected. If the relative is still better, it is maintained within the population.

Once the new population is installed, the process of mutation, recombination and selection is repeated until the optimum is located, or a prespecified termination criterion is satisfied, e.g., the number of generations reaches a preset maximum, gmax [4].

C) Particle Swarm Optimization (PSO): PSO (Particle Swarm Optimization) was proposed by James Kennedy and Russell Eberhart in 1995. It is motivated by social behavior of organisms such as bird flocking and fish schooling [29]. It can be used for nonlinear and mixed integer optimization. PSO is different from evolutionary computing, as in it flying potential solutions through hyperspace are accelerating towards "better" solutions, while in evolutionary computation schemes operate directly on potential solutions which are represented as locations in hyperspace [4]. The position of a particle is influenced by the best position visited by itself (i.e. its own experience) and the position of the best particle in its neighborhood (i.e. the experience of neighboring particles) [30]. Particle position, $x_i$, are adjusted using:

$$x_i(t+1)=x_i(t)+v_i(t+1) \quad (4)$$

where the velocity component, vi, represents the step size. For the basic PSO,

$$v_{i,j}(t+1)=wv_{i,j}(t)+c_1r_{1,j}(t)(y_{i,j}(t)-x_{i,j}(t))+c_2r_{2,j}(t)(\hat{y}_j(t)-x_{i,j}(t)) \quad (5)$$

where w is the inertia weight [31], c1 and c2 are the acceleration coefficients, $r_{1,j}$, $r_{2,j}$ ~ U(0, 1), $y_i$ is the personal best position of particle i, and $\hat{y}_i$ is the neighborhood best position of particle i [30]. The neighborhood best position $\hat{y}_i$, of particle i depends on the neighborhood topology used [32,33].

The main steps involved in PSO are [34]:

1) Initialize a population array of particles with random positions and velocities on D dimensions in the search space.

2) For each particle, evaluate the desired optimization fitness function in D variables.

3) Compare particle's fitness evaluation with its previous best. If current value is better than previous best, then set previous best equal to the current value, and previous best position equal to the current location in D-dimensional space.

4) Identify the particle in the neighborhood with the best success so far.

5) Change the velocity and position of the particle according to (4) and (5)

6) If a criterion is met (usually a sufficiently good fitness or a maximum number of iterations) then optimal result is given out otherwise optimization continues.

DE and PSO have been used to optimize the parameters of SVM during training and then those parameters have been used to create the best possible model for prediction purposes.

## III. Implementation Details

*1) Dataset:* The daily datasets of Honeywell International Inc. (listed on NYSE) and Apple Inc. (listed on NASDAQ), have been used for implementation purposes. The data sets are available at (http://wikiposit.org/Finance/Stocks/) and are available in csv, html, tab delimited, xml and raw formats. The reference site for this data is www.finance.google.com. Opening price, high, low, adjusted closing price and volume have been used as inputs and the closing price the following day is the output for SVM model. The training datasets have 500 records each from 6 April, 2009 to 29 March, 2011 for Honeywell and from 17 July, 2009 to 12 July, 2011 for Apple. The testing datasets have 200 records each from 30 March, 2011 to 12 Jan, 2012 for Honeywell and 13 July, 2011 to 27 April, 2012 for Apple.

The paper compares prediction results of both normalized and non-normalized datasets.

The data has been normalized as inspired by [18] to:

1) Avoid the data with large range "submerge" those with small range and balance their functions in the training to make data comparable [18].
2) To enhance training efficiency and to avoid the problem of inner product calculation when calculating kernel function [18].

The formula used for normalization is [18]:

$$x' = x_{low} + \frac{(x_{up}-x_{low})(x-x_{min})}{x_{max}-x_{min}} \quad (6)$$

Here, x is the original data, x' is the data after normalization, $x_{min}$ is the minimum of original data, $x_{max}$ is the maximum of original data, $x_{low}$ is the lower bound of the data after normalization, $x_{up}$ is the upper bound of the data after normalization. Here, we use $x_{low} = -1$ and $x_{up} = +1$ .

*2) Performance indicator:* The performance measure used is MSE (Mean Square Error):

$$\text{MSE}= \frac{1}{n}\sum_{i=1}^{i=n}(a_i - p_i)^2 \quad (7)$$

Here, a is the actual value, p is the predicted value, i represents the term index which ranges from 1 to n, where n represents the last term index. MSE helps to avoid NAs and negative terms in result which can arise because of normalization of data.

3) Methodology: The basic methodology for the both normalized and non- normalized approaches is same.

To find the optimal range of all three parameters C,ϵ,γ first two parameters are fixed and the other one is varied to see its effect on Training MSE, Testing MSE and number of support vectors. And then the second parameter is fixed and so on. All these collected values are considered to find the optimal range to be used for the purpose of stock price prediction. The general range of these parameters can vary over a large solution space but the optimal range differs for different applications and is also dataset dependent. Training MSE, testing MSE and number of support vectors of all three parameters are checked for overfitting and underfitting to select the optimal range.

The following points have been considered while selecting values of C and γ:

i) Selecting C: A 'good' value for C can be chosen equal to the range of output (response) values of training data [19]. However, such a selection of C is quite sensitive to possible outliers (in the training data) [20] so, C has been fixed using the formula suggested in [20]:

$$C=max(|y'+3\sigma_y|,|y'-3\sigma_y|) \quad (8)$$

Here, y' and $\sigma_y$ are the mean and standard deviation of the y values of the training data. This C value coincides with prescription suggested by Mattera and Haykin (1999) when the data has no outliers, but yields better C-values when the data contains outliers [20]. Based on above formula C is calculated as 69.167.

ii) Selecting γ: RBF kernel has been used for implementation of SVM. This use is inspired from [18]. Radial basis kernel expression is as follows:

$$K(x_i,x)=\exp\left(-\frac{\|x_i-x\|^2}{2\sigma^2}\right) \quad (9)$$

According to [20] for multivariate d-dimensional problems the RBF width parameter should be such that $\sigma^d \sim (0.1\text{-}0.5)$ so γ or $2\sigma^2$ has been selected as 0.0625.

iii) Mattera and Haykin (1999) propose to choose ϵ-value so that the percentage of SVs in the SVM regression model is around 50% of the number of samples [19]. [20] suggests that optimal generalization performances can be achieved with the number of SVs more or less than 50%. The range of values where number of SVs is from 200 to 300 has been chosen for optimization purpose in the implementation.

Dataset for Apple:

Finding range for ϵ:

i) Selecting C: The value of C has been fixed at 450.8346 using (8).

ii) Selecting γ: Value of γ is fixed at 0.0625 according to [20] as explained above.

i) Normalized dataset parameters decision making:

*Finding range for ϵ:* After fixing values of C and γ at 450.8346 and 0.0625 respectively, the values of different aspects for ϵ have been calculated over the range [0.01,0.30]. The results for no. of support vectors, training MSE and testing MSE are shown in Figure 1(a), 1(b) and 1(c) respectively. The favorable range for ϵ has been found as [0.033,0.052] based on required number of support vectors, decrease in training and testing MSEs.

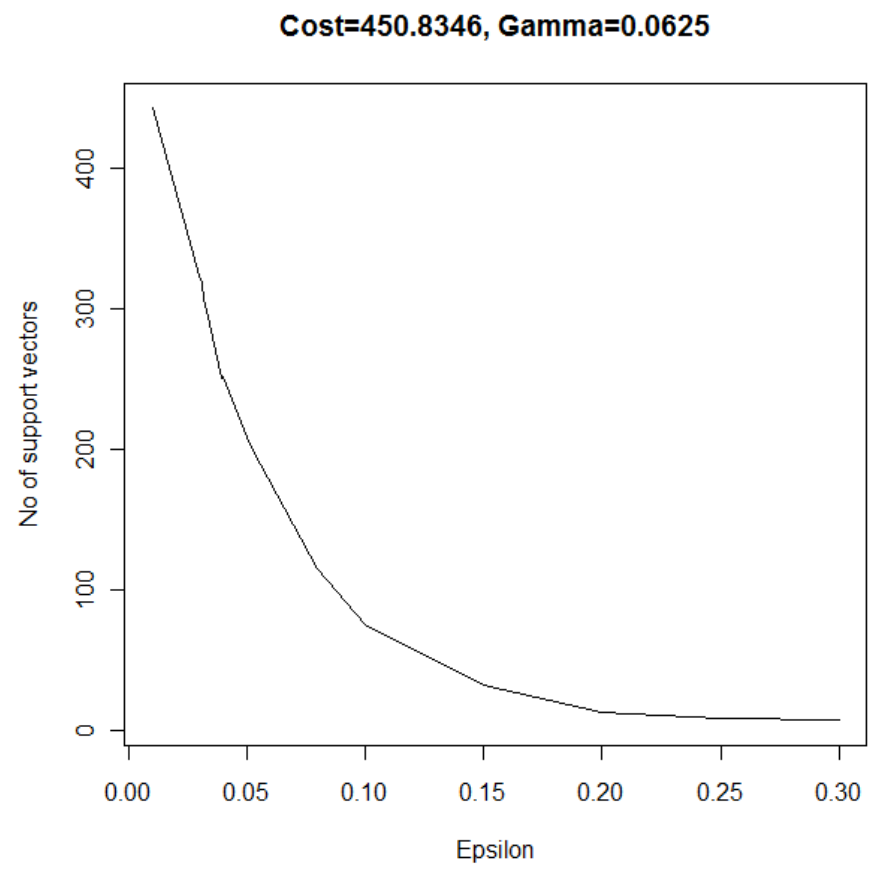

Figure 1(a)

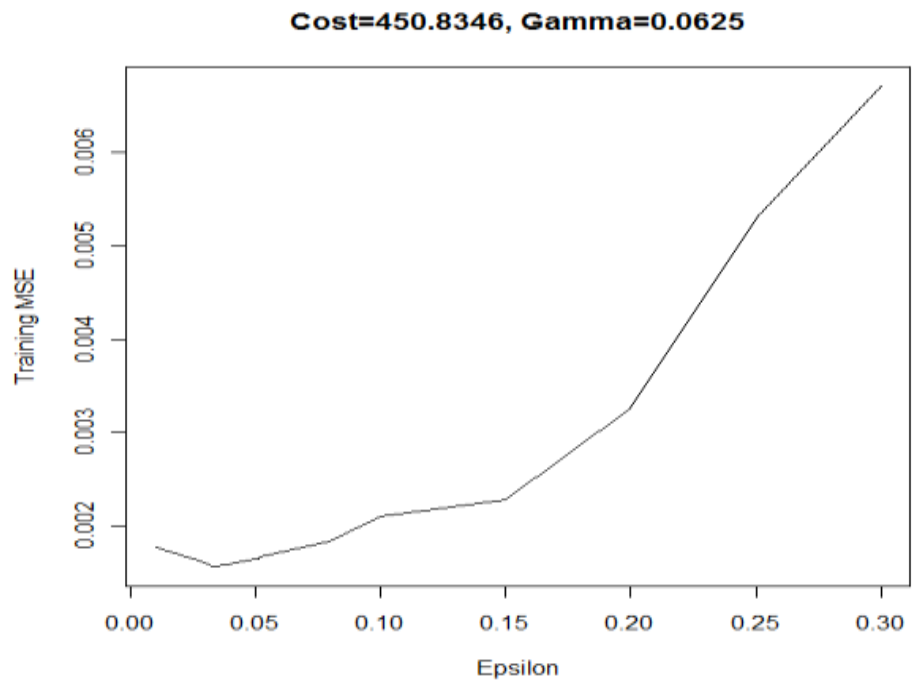

Figure 1(b)

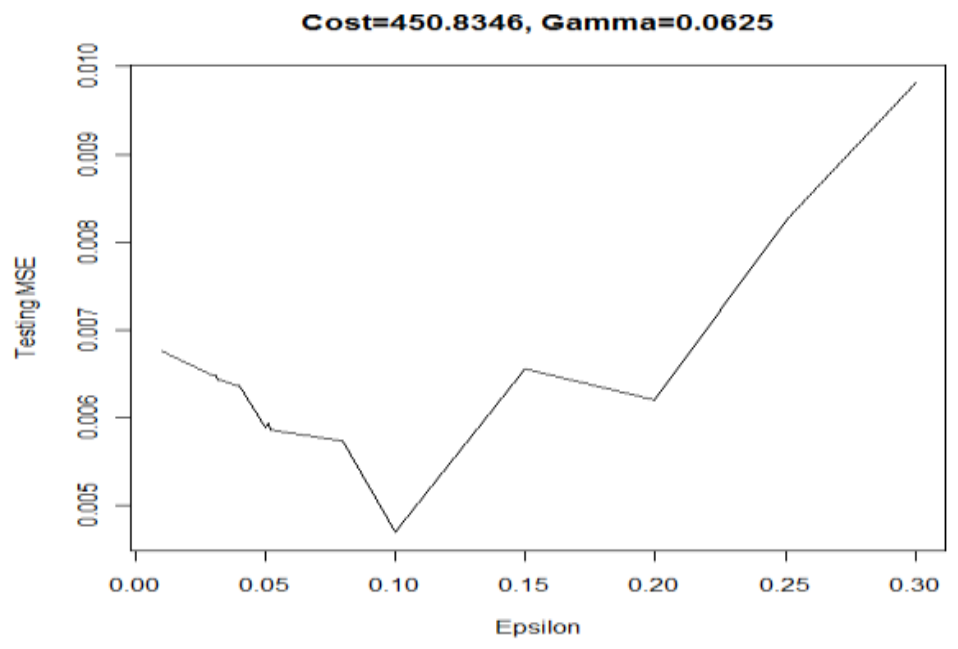

Figure 1(c)

*Finding range for C:* i) ϵ has been selected from above found range of [0.033,0.052]. It has been set as 0.039.
ii) γ is set as 0.0625.

The values of C are examined over [0.1,6000] while fixing ϵ and γ. The results of no of support vectors, training and testing errors are shown in Figure 2(a), 2(b) and 2(c). The range of C has been selected as [1,550]. Figure 2(a) shows that number of support vectors never fall below 200. So, C has been selected such that training MSE decreases and there is no significant increase in testing MSE.

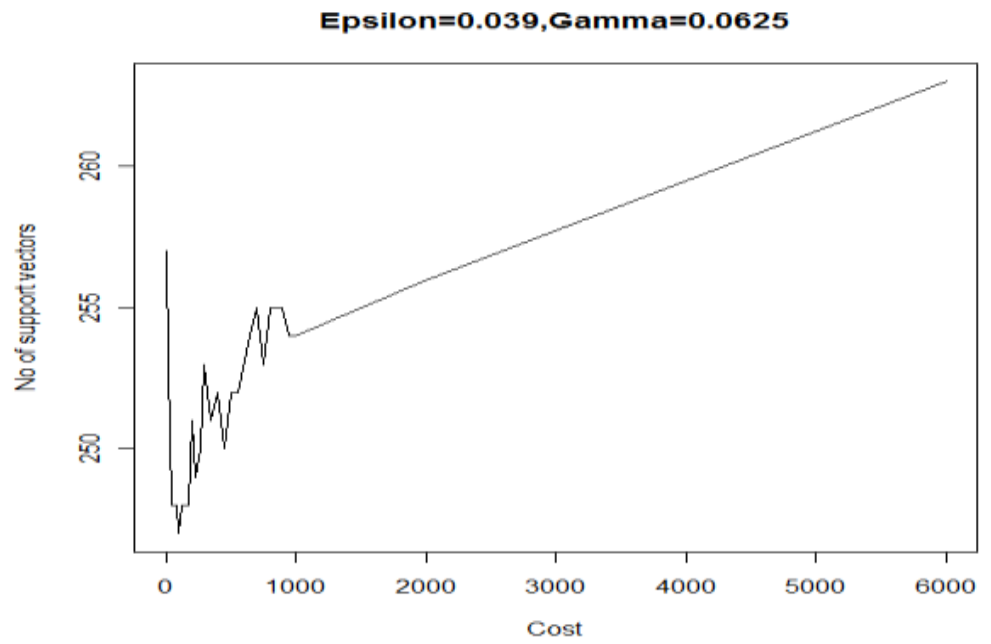

Figure 2(a)

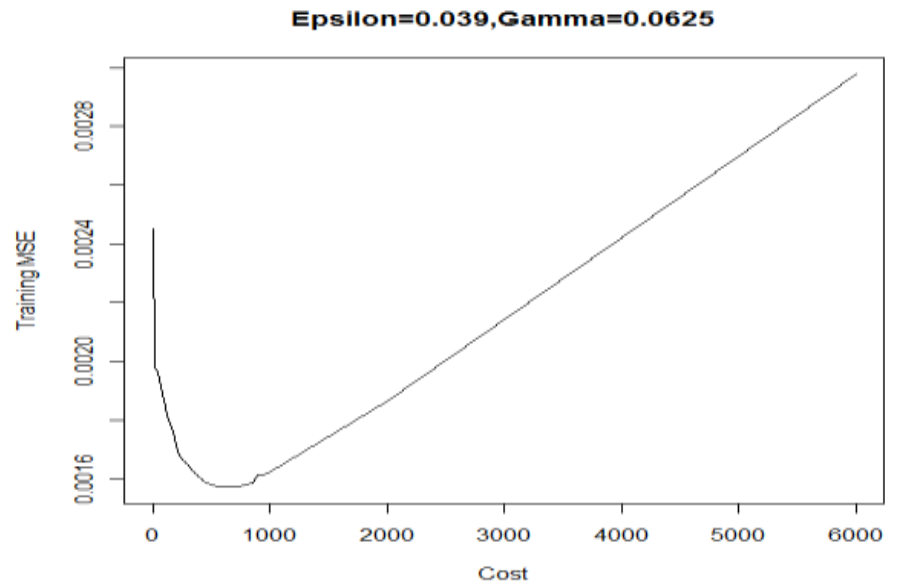

Figure 2(b)

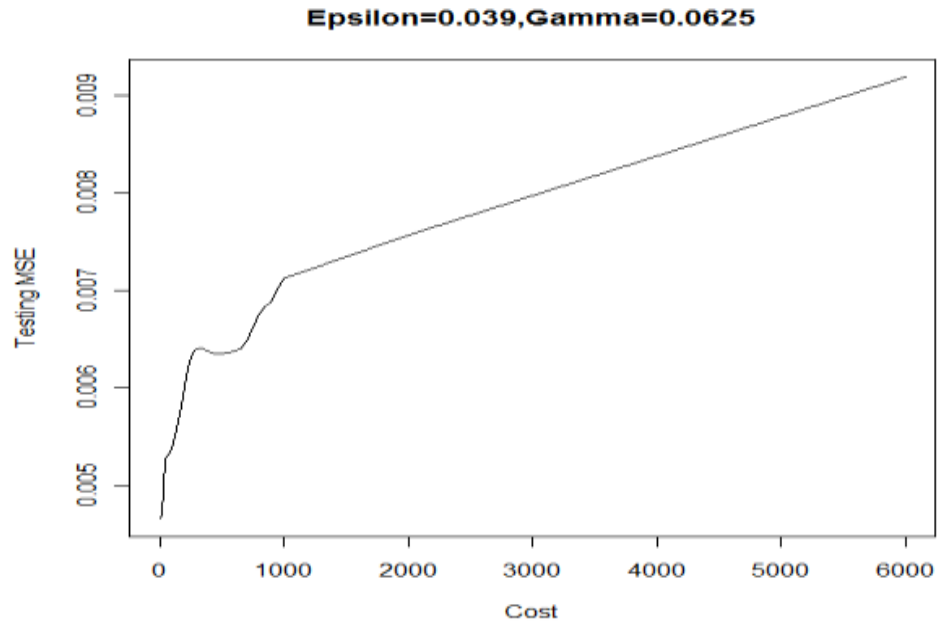

Figure 2(c)

*Finding range for γ:* i) ) ϵ has been selected from above found range of [0.033,0.052]. It has been set as 0.039.

ii) C has been selected from above found range of [1,550]. It has been set as 500.

The values of γ are examined over [0.0,0.4] after fixing ϵ and C. The results for no of support vectors, training and testing MSEs are shown in Figure 3(a), 3(b) and 3(c). The range of γ has been selected as [0.01,0.11]. Figure 3(a) shows that number of support vectors never fall below 200. The range has been selected so that there is no significant increase in training and testing MSEs.

*ii) Non-normalized dataset parameters decision making:*

*Finding range for ϵ:* After fixing the values of C and γ at 450.8346 and 0.0625, the SVM model is created, training and testing MSEs along with no. of support vectors are recorded for ϵ over the range of [0.01,0.20].

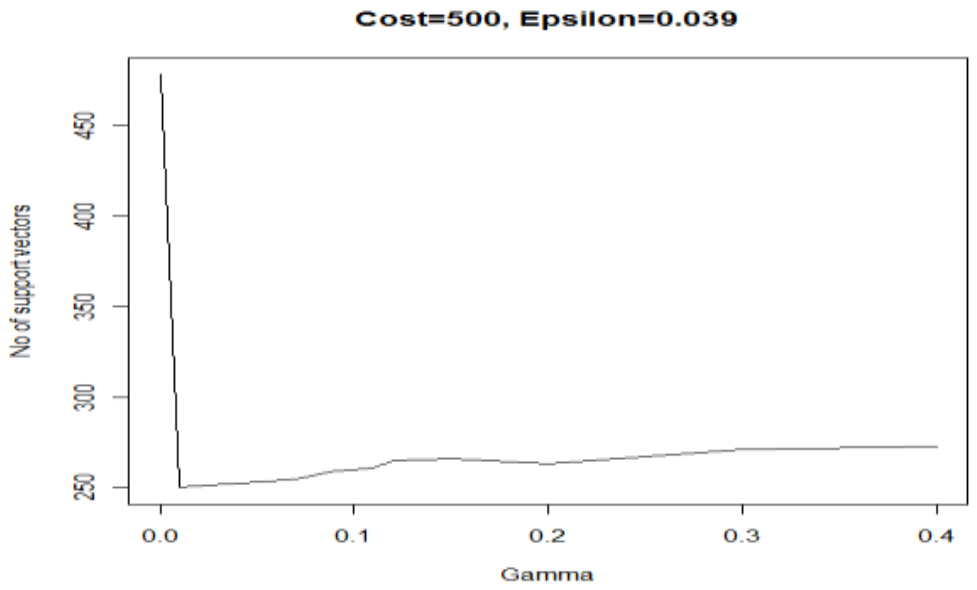

Figure 3(a)

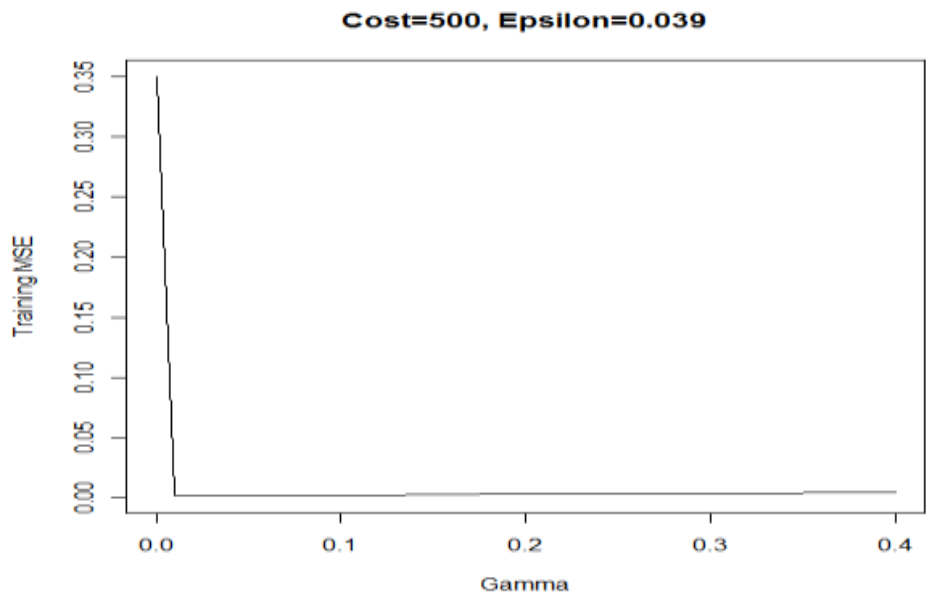

Figure 3(b)

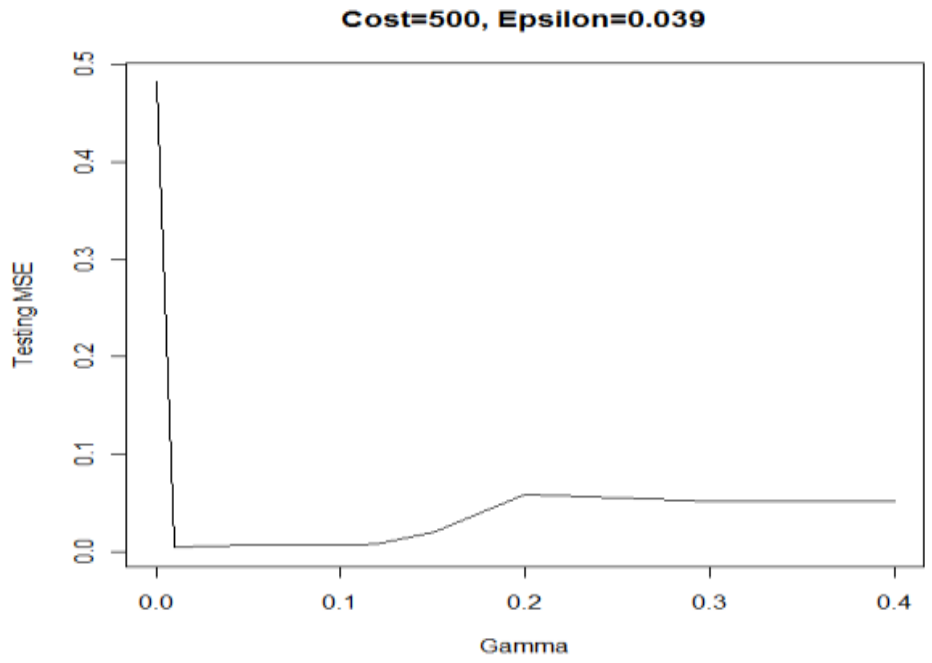

Figure 3(c)

The results are shown in Figures 4(a), 4(b) and 4(c). The range for ϵ has been selected as [0.033,0.052] after considering appropriate number of support vectors and after examining that training and testing errors don’t increase significantly in this range.

*iii) Finding range for C:*

i) ϵ has been selected from above found range of [0.033,0.052]. It has been set as 0.035.

ii) γ is set as 0.0625.

The results for no of support vectors, training error and testing error over range of C~[1,3000] are shown in Figures 5(a), 5(b) and 5(c).

The range for C has been selected as [1,300].

Figure 5(a) shows that number of support vectors never fall below 200. The range has been selected such that training error decreases.

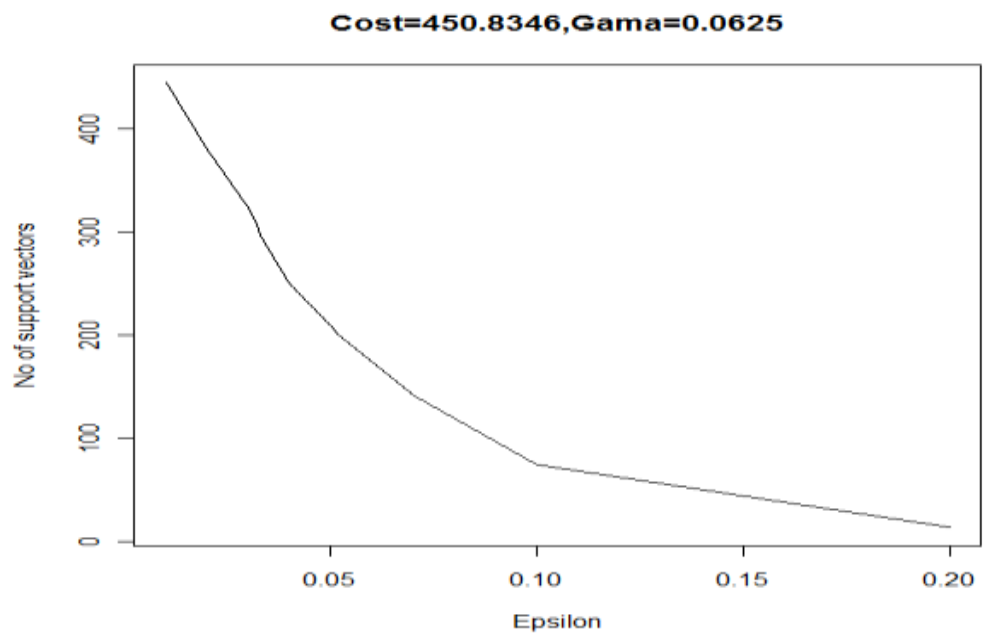

Figure 4(a)

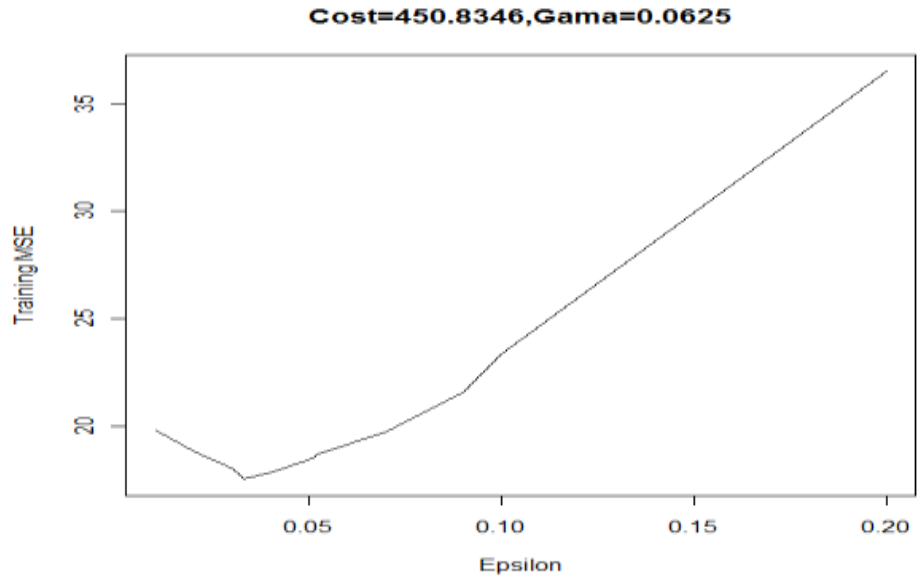

Figure 4(b)

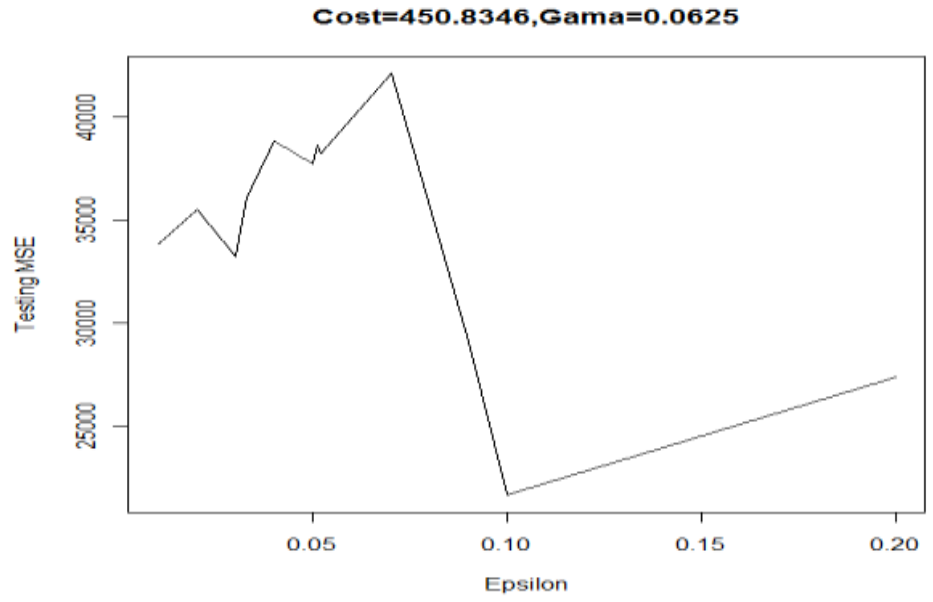

Figure 4(c)

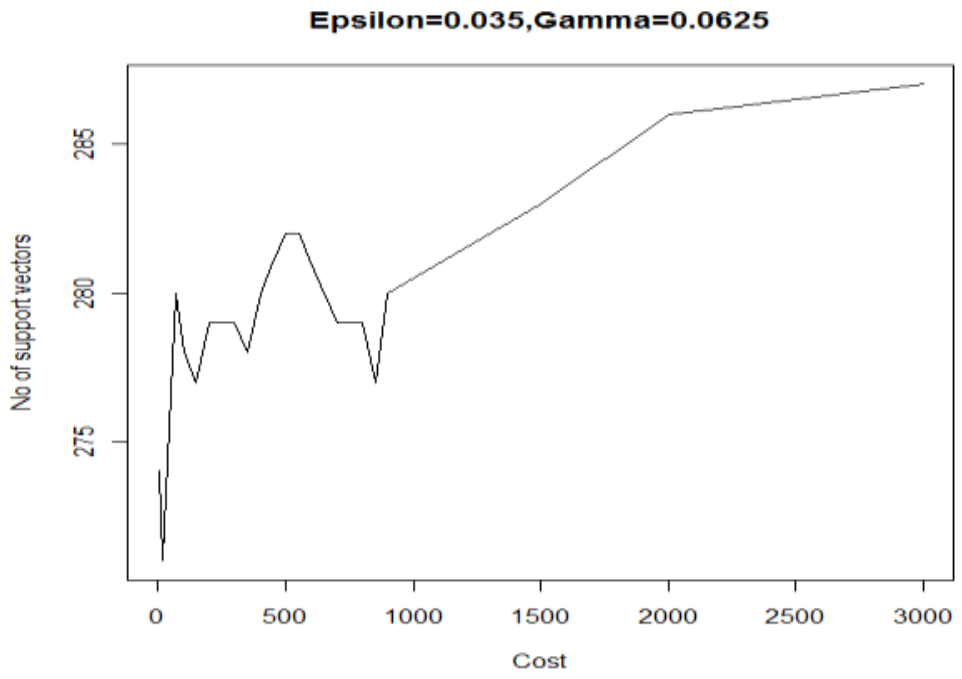

Figure 5(a)

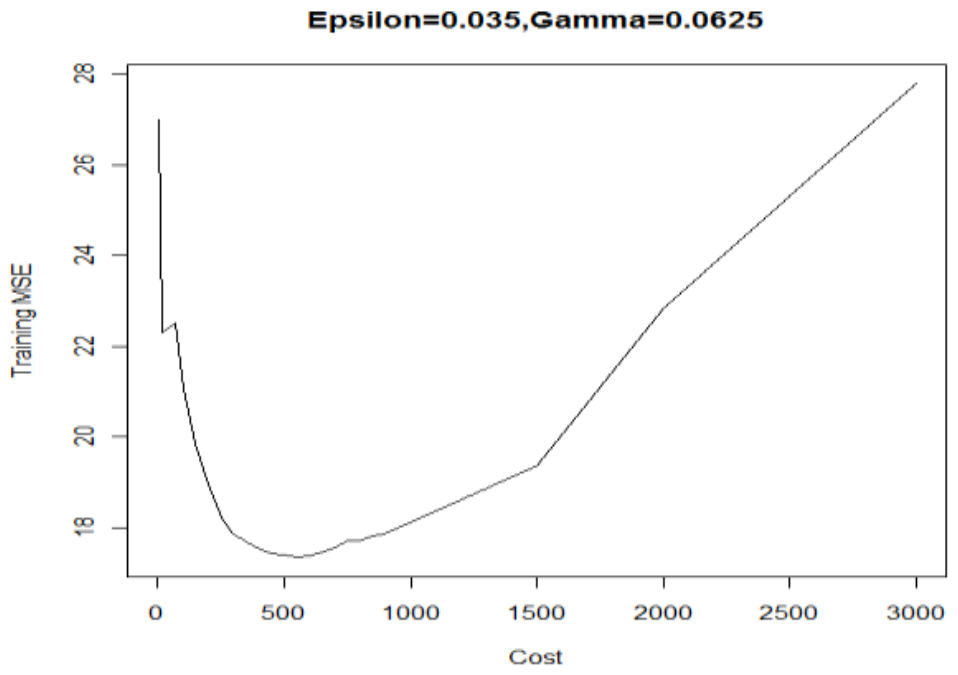

Figure 5(b)

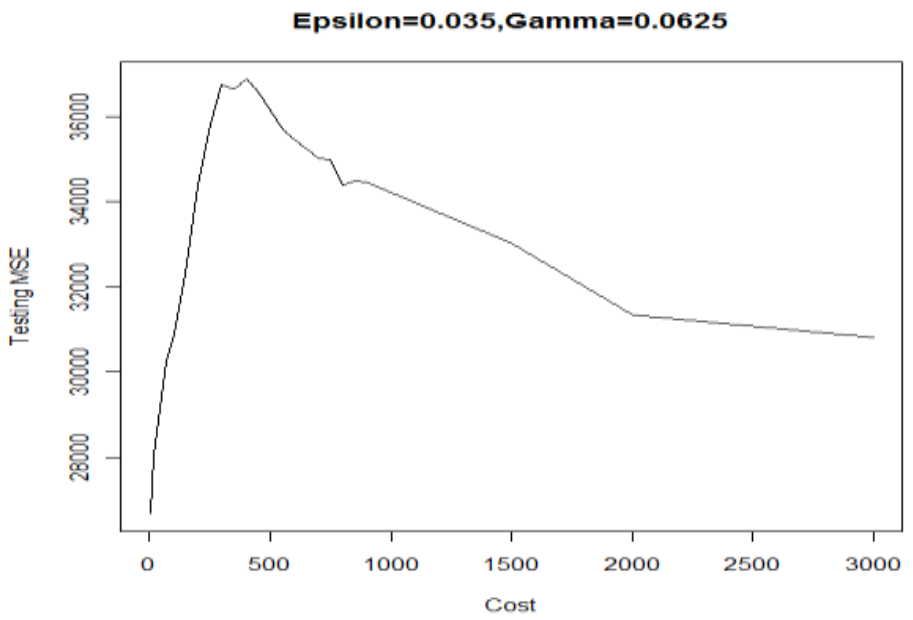

Figure 5(c)

*Finding range for γ:*

i) ) ϵ has been selected from above found range of [0.033,0.052]. It has been set as 0.035.

ii) C has been selected from above found range of [1,300]. It has been set as 200.

γ has been examined over [0.0,0.4] and the results are shown in Figure 6(a), 6(b) and 6(c). The range of γ has been selected as [0.01,0.1]. At γ > 0.1 testing MSE decreases but training error increases.

*Dataset for Honeywell:* The above approach was also used with Honeywell dataset, both normalized and non normalized.

*i) Normalized dataset:* For ϵ, C and γ were fixed at 69.167 and 0.0625 using (8) and according to [20] which gave range as [0.08, 0.15]. Setting ϵ at 0.1 from the selected range and γ at 0.0625, C has favorable range in [1,440]. Now, ϵ at 0.1 and C at 210 from selected favorable range γ had favorable range in [0.02,0.08].

*ii) Non normalized dataset:* For ϵ, C and γ were fixed at 69.167 and 0.0625 using (8) and according to [20] which gave range as [0.05,0.07]. Setting ϵ at 0.05 from the selected range and γ at 0.0625, C has favorable range in [1,60]. Now, ϵ at 0.05 and C at 30 from selected favorable range γ had favorable range in [0.01,0.1].

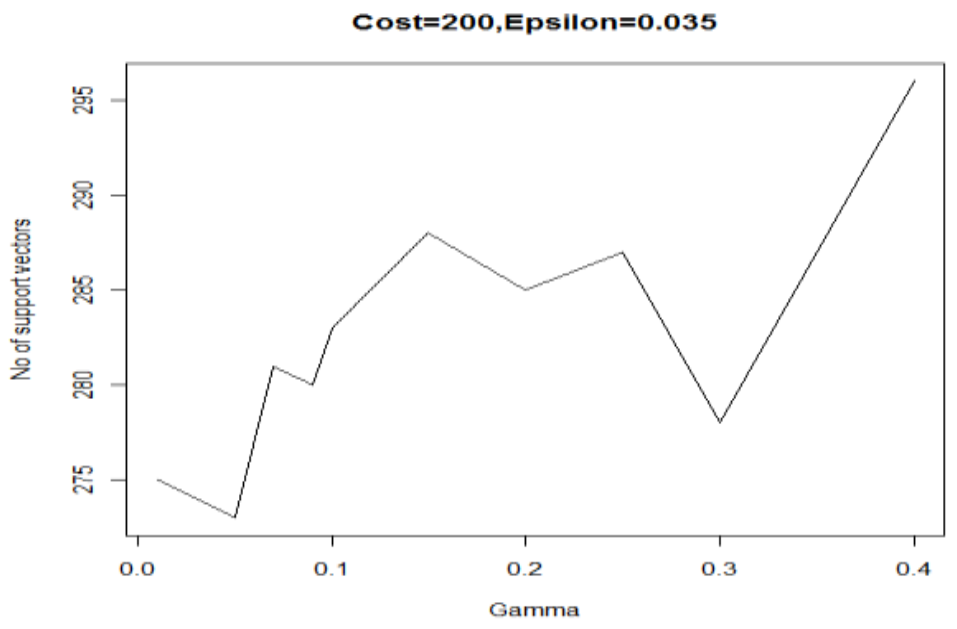

Figure 6(a)

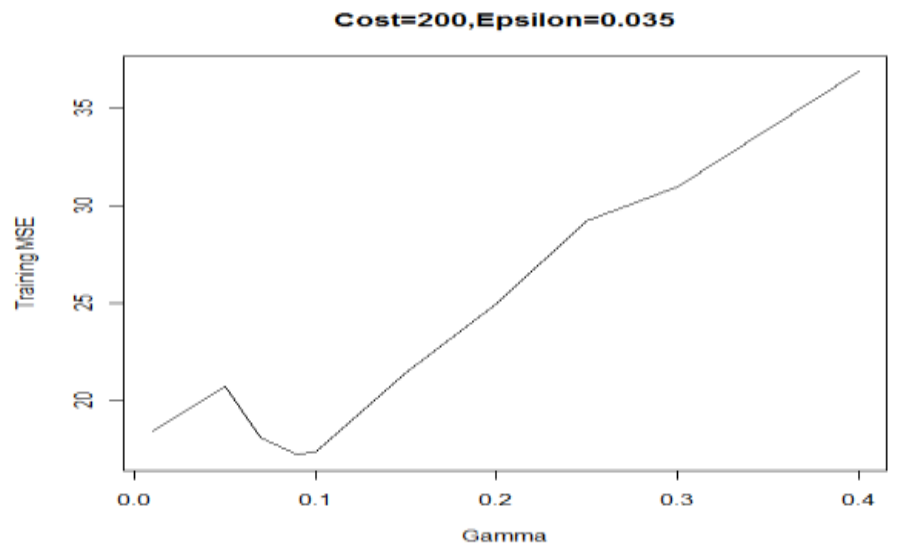

Figure 6(b)

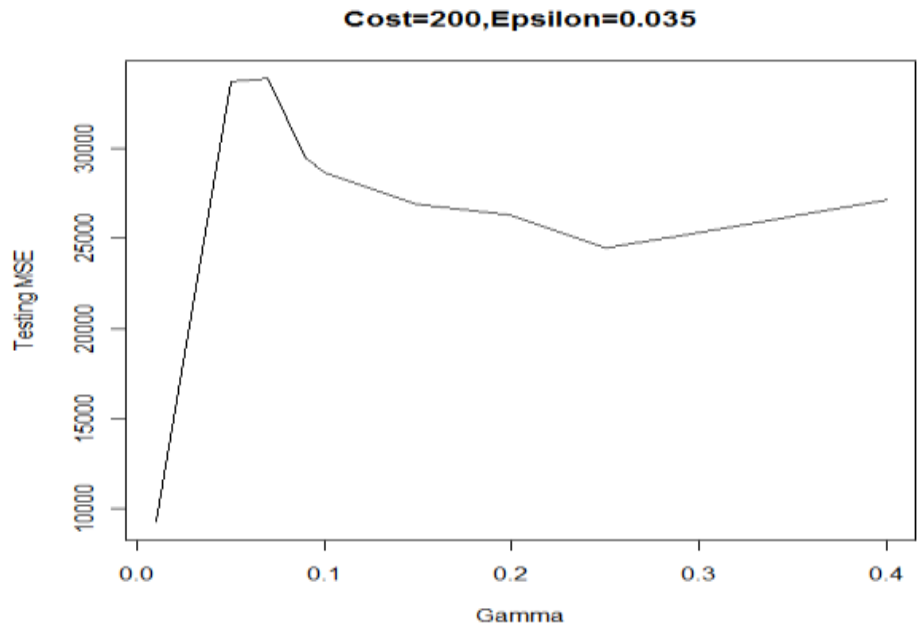

Figure 6(c)

4) DE-SVM model:

All implementation has been done in R on a system with AMD Turion-X2 2GHz Dual Core processor having 2GB RAM and Windows 7 Ultimate (32 bit) OS. The model used is shown in Figure 7.

## IV. Results

Apple:

*Normalized dataset:* The range of parameters C,ϵ,γ are [1,550], [0.033,0.052] and [0.01,0.11] respectively. The time taken for both PSO and DE to converge is 13hrs approx. The results for both DE and PSO are shown in Table 1. Table 2 shows prediction results for SVM (with default parameters of C=1, ϵ=0.1,γ=0.2), DE-SVM and PSO-SVM together.

*Non normalized dataset:* The range of parameters C,ϵ,γ are [1,300], [0.033,0.052] and [0.01,0.1] respectively. The results for DE-SVM and PSO-SVM are shown in Table 3. Table 4 shows prediction results for SVM (with default parameters of C=1, ϵ=0.1,γ=0.2), DE-SVM and PSO-SVM together. The large values of MSEs for testing are because of the highly inaccurate predicted values produced because of the wide range of the output values.

Table 1

| | Optimized C,ϵ,γ | Training MSE | Testing MSE | No of support vectors |
|---|---|---|---|---|
| DE | 286.37295110, 0.03567755, 0.08290609 | 0.001520 | 0.006520839 | 277 |
| PSO | 312.57590986, 0.03556743, 0.08097982 | 0.001519326 | 0.006537451 | 277 |

Table 2

| | Training MSE | Testing MSE | No. of support vectors |
|---|---|---|---|
| SVM | 0.003224442 | 0.008572274 | 94 |
| DE-SVM | 0.001520 | 0.006520839 | 277 |
| PSO-SVM | 0.001519326 | 0.006537451 | 277 |

For both DE and PSO, normalized and non normalized cases, the population size has been fixed at 30 and iterations at 200. The prediction results of DE-SVM and PSO-SVM are better than SVM alone in both cases.

Table 5 shows predicted stock price values for both normalized and unnormalized datasets for SVM, DE-SVM and PSO-SVM models.

Table 3

| | Optimized C,ϵ,γ | Training MSE | Testing MSE | No. of support vectors |
|---|---|---|---|---|
| DE | 298.574181, 0.035675, 0.082388 | 17.029155 | 30013.67 | 276 |
| PSO | 296.05980293, 0.03569455, 0.08201110 | 17.01084 | 30175.51 | 276 |

Table 4

| | Training MSE | Testing MSE | No. of support vectors |
|---|---|---|---|
| SVM | 36.10602 | 30383.61 | 94 |
| DE-SVM | 17.029155 | 30013.67 | 276 |
| PSO-SVM | 17.01084 | 30175.51 | 276 |

Table 5

| Original | Normalized data | | | Unnormalized data | | |
|---|---|---|---|---|---|---|
| price | SVM | DE-SVM | PSO-SVM | SVM | DE-SVM | PSO-SVM |
| 603 | 623.7251 | 615.6471 | 615.6017 | 264.1839 | 248.7355 | 248.1471 |
| 545.17 | 552.267 | 553.2582 | 553.4743 | 264.2948 | 263.0166 | 262.1655 |
| 493.42 | 497.0696 | 496.938 | 496.959 | 267.2491 | 264.4146 | 263.1924 |
| 369.8 | 359.912 | 367.1551 | 367.1906 | 354.1847 | 359.2065 | 359.2020 |

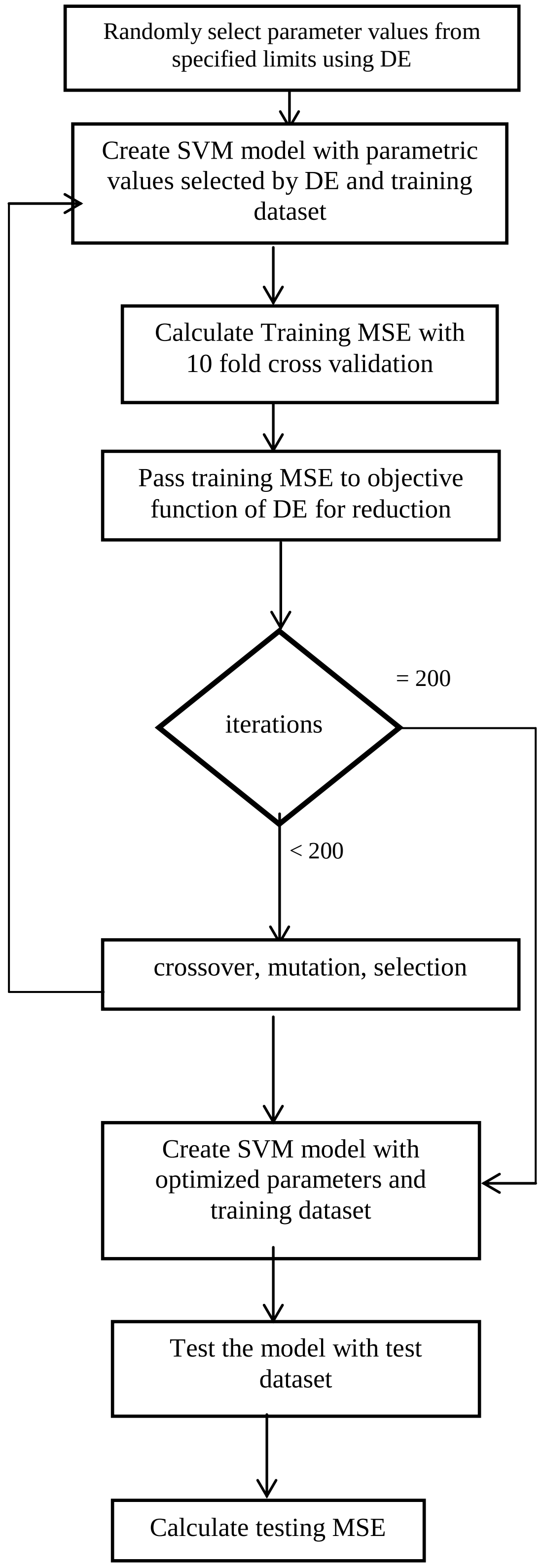

Figure 7

Honeywell:

Normalized dataset: The range of parameters C,ϵ,γ are [1,440],[0.08,0.15] and [0.02,0.08] respectively. Table 6 shows prediction results.

Table 6

| SVM | Optimized C, ϵ,γ | Training MSE | Testing MSE | No. of support vectors | Time taken |
|---|---|---|---|---|---|
| DE (CR= 0.7, F= 0.9) | 439.864990, 0.080024, 0.079993 | 0.003211084 | 0.03108918 | 222 | 4 hrs 10 min |
| PSO | 440, 0.08 0.07999401 | 0.003210875 | 0.03117105 | 222 | 4 hrs 30 min |

Non normalized dataset results: The range of parameters C,ϵ,γ are [1,60],[0.05,0.07] and [0.01,0.1] respectively. Table 7 shows the results. Predicted values for both normalized and non normalized datasets are shown in Table 8.

For DE CR=0.7 and F=0.9. DE / local-to-best / 1 / bin strategy has been used for DE for all the implementations in this paper.

Table 7

| | C,ϵ,γ | Training MSE | Testing MSE | No. of support vectors |
|---|---|---|---|---|
| DE-SVM | 40.543474, 0.056122, 0.010113 | 0.4726898 | 1.603333 | 256 |
| PSO-SVM | 40.7422794, 0.06411372, 0.01000007 | 0.4727329 | 1.620285 | 225 |
| SVM | 1,0.1,0.2 | 0.8239745 | 3.8681 | 148 |

Table 8

| Original | Normalized | | | Unnormalized | | |
|---|---|---|---|---|---|---|
| Closing price | SVM | DE-SVM | PSO-SVM | SVM | DE-SVM | PSO-SVM |
| 41.94 | 42.503 | 42.340 | 42.34 | 43.019 | 42.95 | 42.95808 |
| 62 | 61.555 | 61.63 | 61.64 | 56.079 | 61.56 | 61.53316 |
| 43.22 | 44.15 | 44.002 | 44.00 | 45.36 | 44.65 | 44.66156 |
| 53.56 | 53.84 | 54.15 | 54.15 | 54.185 | 54.44 | 54.47695 |
| 56.43 | 56.373 | 56.64 | 56.65 | 56.546 | 56.654 | 56.70812 |

## V. CONCLUSION

The performance of SVM can be significantly affected by choice of its free parameters of cost (C), insensitive loss function (ϵ) and kernel parameter (γ). The results show that DE-SVM model's performance is comparable to that of PSO-SVM. Performance of these models can be improved by normalization of datasets. Normalization helps to significantly improve the accuracy of the output when the range of values is vast. Normalization gives equal weightage to all the input variables by converting the values of all the variables within a pre-specified range. This helps to avoid dominance of one variable over others in the created model. So, it helps to improve the efficiency of the created model. SVM alone performs better when data is normalized because in hybrid models optimization techniques help to tune the model according to requirement of datasets. With normalization of data, the range for optimization of C,ϵ,γ improves.

## VI. Future Scope

DE responds to the population progress after a time lag. The whole population in DE remains unchanged until it is replaced by a new population [15]. Hence, it results in slower convergence. To alleviate this problem, a dynamic version of DE called Dynamic Differential Evolution (DDE) has been proposed by Anyong Qing [23]. DEPSO algorithm, which represents more stability by dual evolution, proposed by Ying-Chih Wu [24] can be used for optimization of SVM. The above mentioned methods will help to further improve the efficiency of SVM and hence improve results.